\documentclass[sigconf]{acmart}

\usepackage{balance}

\def\BibTeX{{\rm B\kern-.05em{\sc i\kern-.025em b}\kern-.08emT\kern-.1667em\lower.7ex\hbox{E}\kern-.125emX}}
    
\copyrightyear{2019}
\acmYear{2019}
\setcopyright{acmcopyright}
\acmConference[MM '19] {Proceedings of the 27th ACM International Conference on Multimedia}{October 21--25, 2019}{Nice, France}
\acmBooktitle{Proceedings of the 27th ACM International Conference on Multimedia (MM'19), Oct. 21--25, 2019, Nice, France}
\acmPrice{15.00}
\acmDOI{10.1145/3343031.3350995}
\acmISBN{978-1-4503-6889-6/19/10}

\usepackage{algorithm}  
\usepackage{algorithmicx}  
\usepackage{algpseudocode}  
\usepackage{amsmath}  
\setcopyright{acmcopyright}

\copyrightyear{2019} 
\acmYear{2019} 
\acmConference[MM '19]{Proceedings of the 27th ACM International Conference on Multimedia}{October 21--25, 2019}{Nice, France}
\acmBooktitle{Proceedings of the 27th ACM International Conference on Multimedia (MM '19), October 21--25, 2019, Nice, France}
\acmPrice{15.00}
\acmDOI{10.1145/3343031.3350995}
\acmISBN{978-1-4503-6889-6/19/10}

\usepackage{setspace}
\begin{document}
\fancyhead{}
%
\title{Adaptive Semantic-Visual Tree for Hierarchical Embeddings}

\author{%
    Shuo Yang $^{1,2*}$,
    Wei Yu $^2$,
    Ying Zheng $^1$,
    Hongxun Yao $^{1*}$,
    Tao Mei $^2$
}

\affiliation{%
$^1$Harbin Institute of Technology \\
$^2$JD AI Research \\
}
\email{{syang, zhengying, h.yao}@hit.edu.cn,{yuwei10,tmei}@jd.com}

\begin{abstract}
Merchandise  categories inherently form a semantic hierarchy with different levels of concept abstraction, especially for fine-grained categories. This hierarchy encodes rich correlations among various categories across different levels, which can effectively regularize the semantic space and thus make prediction less ambiguous. However, previous studies of fine-grained image retrieval primarily focus on semantic similarities or visual similarities. In real application, merely using visual similarity may not satisfy the need of consumers to search merchandise with real-life images, e.g., given a red coat as query image, we might get red suit in recall results only based on visual similarity, since they are visually similar; But the users actually want coat rather than suit even the coat is with different color or texture attributes. We introduce this new problem based on photo shopping in real practice. That's why semantic information are integrated to regularize the margins to make "semantic" prior to "visual". To solve this new problem, we propose a hierarchical adaptive semantic-visual tree (ASVT) to depict the architecture of merchandise categories, which evaluates semantic similarities between different semantic levels and visual similarities within the same semantic class simultaneously. The semantic information satisfies the demand of consumers for similar merchandise with the query while the visual information optimize the correlations within the semantic class. At each level, we set different margins based on the semantic hierarchy and incorporate them as prior information to learn a fine-grained feature embedding.  To evaluate our framework, we propose a new dataset named JDProduct,  with hierarchical labels collected from actual image queries and official merchandise images on online shopping application. Extensive experimental results on the public CARS196 and CUB-200-2011 datasets demonstrate the superiority of our ASVT framework against compared state-of-the-art methods.
\end{abstract}

%
%

%

%

%
\maketitle
\begin{small}
\begin{spacing}{1}
\textbf{ACM Reference Format:} \\
Shuo Yang, Wei Yu, Ying Zheng, Hongxun Yao, Tao Mei. 2019. Adaptive Semantic-Visual Tree for Hierarchical Embeddings. In \emph{Proceedings of the 27th ACM International Conference on Multimedia (MM'19), Oct. 21--25, 2019, Nice, France.} ACM, New York, NY, USA, 9 pages. https://doi.org/10.1145/3343031. 3350995
\end{spacing}
\end{small}
\section{Introduction}
\begin{figure}[t]
\label{fig_firstpage}
\centering
    \includegraphics[width=1\linewidth]{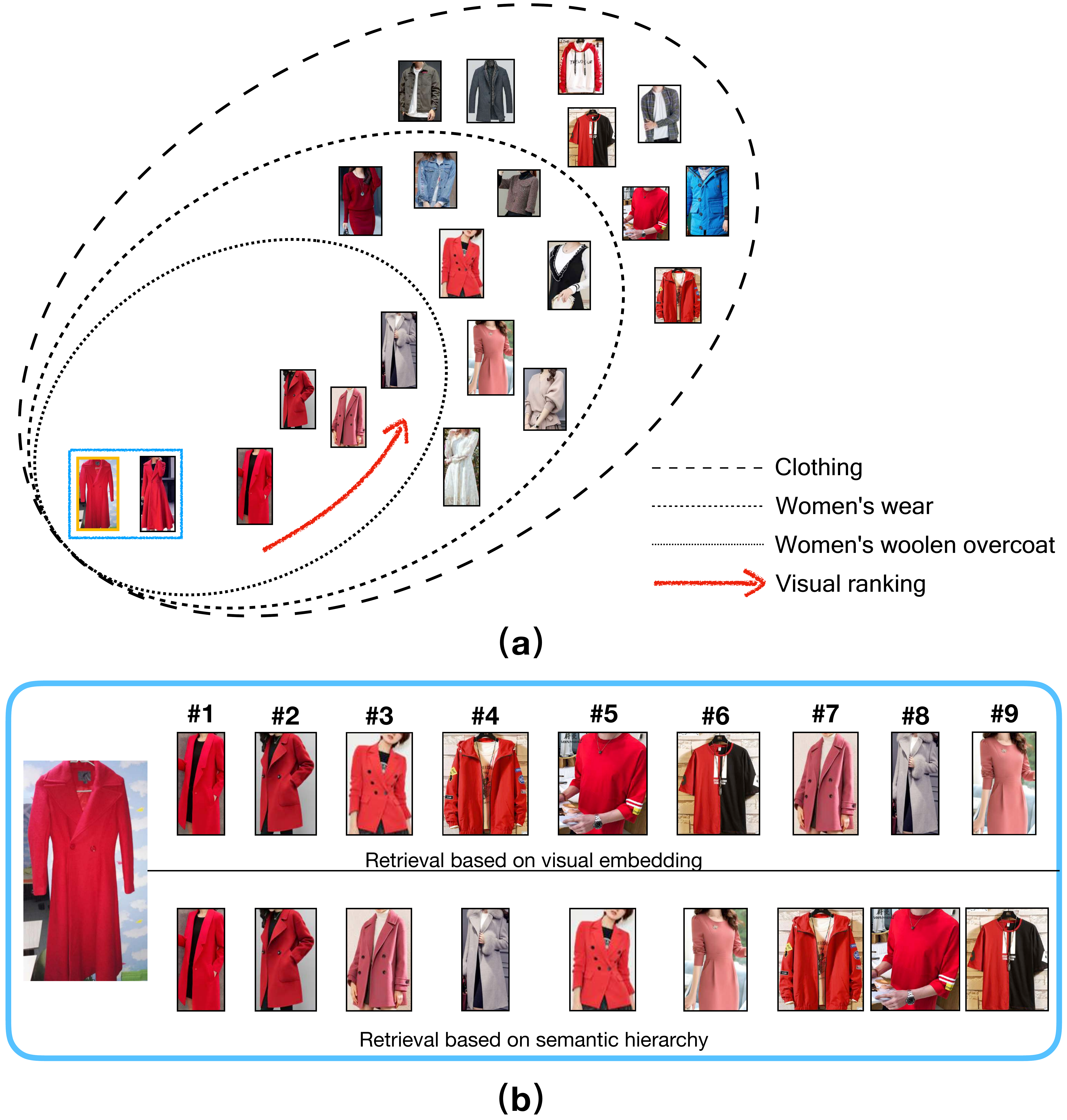}
\caption{(a) The actual demand of image shopping. Merchandise inherently have multiple labels with different levels of semantic abstraction. Users want the results to be from exactly the same or similar semantic class, which is prior to visual similarity.(b) The comparison of retrieval results of women's red overcoat between visual embedding and semantic hierarchy. With only visual embedding, men's red clothes would also be retrieved prior to women's gray overcoat as in the first row, while the second row doesn't have this problem.} 
\label{fig_firstpage}
\end{figure}
As one form of electronic commerce, online shopping allows consumers to search and buy merchandise  on shopping platforms like Amazon, Taobao, and JD.com. However, the name of a certain good is unknown to consumers; they need a more convenient method to directly find what they want -- using image search engine.  And this kind of tasks to find the image with a query image can be called content-based image retrieval(CBIR). The core mission of CBIR in online shopping is to find the exactly the same instance, and find the merchandise  from the same category if there is no same instance. Such is the basic requirement for online shopping that the retrieval series of the system should have different priorities for different results instance-level similarity>category-level similarity>dissimilarity as in Fig.1(a).

To measure the similarity between the images from the gallery with the query image, we need a powerful tool -- deep metric learning.

 \begin{figure*}[t]
\centering
    \includegraphics[width=0.8\linewidth]{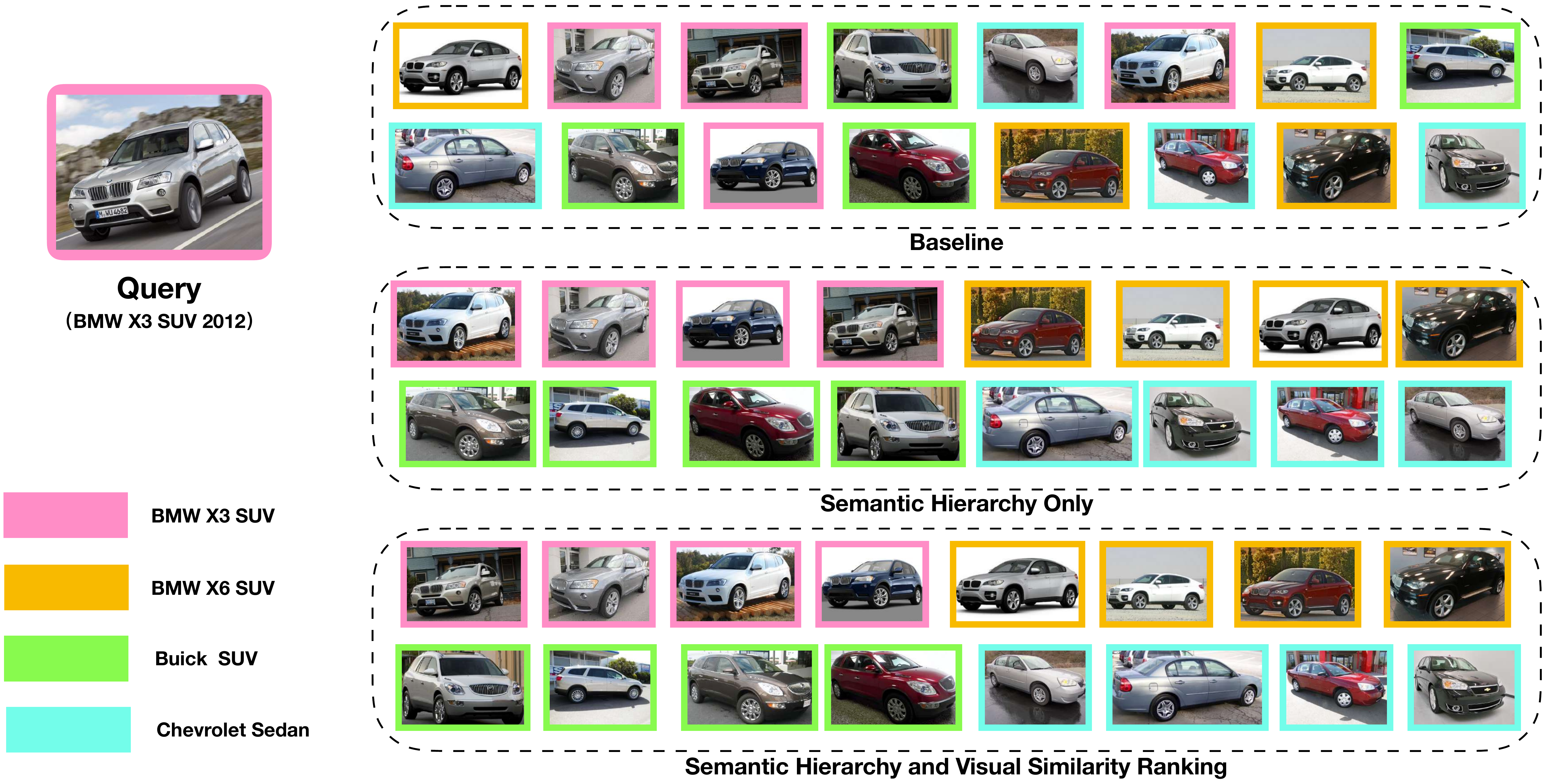}
\caption{The retrieval results of three different methods on CARS196 dataset. The bottom row is the the conventional feature embedding method based on visual similarity. The middle row is feature embedding method based on semantic hierarchy only. And the top shows the results of our proposed adaptive semantic-visual strategy which takes both the semantic hierarchy and visual similarity into consideration. The image with grey box is the query image. To exemplify the necessity of the combination of semantic and visual information in limited space, we manually select 16 images with semantic and visual diversity to form a gallery, and use three models trained from the above methods to compute the distance between the query and the images from the gallery. The retrieval results are ordered from left-top to right-bottom.}
\label{fig_main}
\end{figure*}

Deep metric learning, which advanced the state-of-the-art in various computer vision tasks\cite{DBLP:journals/corr/abs-1801-04815,DBLP:conf/eccv/GeHDS18,DBLP:conf/cvp/SongXJS16}, aims at learning appropriate similarity measurements between pairs of images. And it is a suitable tool to deal with CBIR problems since CBIR needs powerful image descriptors for searching images from a database. There are some instance-level CBIR systems, such as person re-identification\cite{DBLP:journals/tip/DaiZWLW19,DBLP:journals/access/ChenGFXY18,DBLP:journals/pr/LiuMWW18}, or some category-level systems, such as animal recognition\cite{DBLP:conf/dsaa/NguyenMNNFARP17}.

The existed CBIR works using deep metric learning based on online-shopping either only focus on instance-level or category-level, ignoring the fact that online merchandise are usually tagged with multiple levels. Also, there is no such a suitable benchmark offering hierarchical information with multiple levels.

In this paper, we introduce a novel hierarchical adaptive semantic-visual tree (ASVT) to capture the category structure with different levels for online shopping and propose a hierarchical feature embedding method, which measures the instance-level and category-level similarities simultaneously.

Merchandise  categories innately contain a hierarchy with different levels of concept abstraction (e.g. A running shoe is a shoe before it is specified for running). In the semantic tree of merchandise, nodes closer to the root of the hierarchy refer to more abstract concepts while nodes closer to the leaves indicate fine-grained concepts as in Fig.\ref{fig_tree}. But sometimes, just using semantic information may cause the retrieval results to be in the same category but distant visually, while only using visual similarity may cause the retrieval results have merchandise from other categories as in Fig.\ref{fig_firstpage}(b); That's the reason why ASVT is needed to combine the visual information with semantic information. With visual similarities, we could effectively regularize the hypothesis space of functions and provide rich guidance to make adaptive margin selections in the contrastive loss.

For example, when we train our network, we would evaluate the number of matched categories of a contrastive pair. Take two running shoes with different model as example: they are both shoes (so we set margin to be smaller), and they are both running shoes (so we set margin to be even smaller); if four categories are identical we regard them as the same thing but visual information still regularize the margin with visual similarities computed by Euclidean distance. As visual similarities are computed from features extracted by our network, the margins would be changed adaptively while the parameters of the network are updated. Besides, collecting and creating meaningful training samples (e.g., in pairs, triplets or quadruplets) has been a central issue for distance metric learning or similarity learning, and an efficient sampling strategy is of critical importance. To recover the global distribution of training data in a mini-batch, we propose a sampling strategy in our training process.

 To evaluate the effectiveness of our proposed model in actual practice, we collect a new dataset with hierarchical labels collected from actual image queries and official merchandise  images on online shopping application as in Fig\ref{fig_dataset}. This is the first dataset collected from actual user queries and official online merchandise  images which can help many state-of-the-art machine learning techniques fit the actual industrial or commercial demand.Besides, we use CARS196 dataset and CUB-200-2011 dataset to further evaluate our model.

The main contributions of this paper can be concluded in three-fold: \\
1) We formulate a novel hierarchical adaptive semantic-visual tree (ASVT) which incorporates semantically and visually structured information of category hierarchy into the deep neural network for the feature embedding and CBIR. To our knowledge, this is the first work that explicitly combine both structured information to guide CBIR via feature embedding. \\
2) We introduce hierarchical margins into the contrastive loss based on the ASVT tree. \\
3) We collect a new large-scale online merchandise  dataset from JD.com online shopping that covers four-level merchandise  categories for evaluation, named JDProduct dataset. Our proposed dataset is the first that involves in four-level categories coming from real application and they may benefit research on multi-granularity image retrieval.

\section{Related Work}
\subsection{Feature embedding}
Deep metric learning maps an image into a feature vector in a manifold space via common deep neural networks. In this manifold space, Euclidean distance can be directly used as the distance metric to evaluate how far it is between two points. There are many metric learning algorithms, such as \cite{DBLP:conf/cvp/SongXJS16,DBLP:conf/cvpr/SchroffKP15,DBLP:conf/cvpr/ChenCZH17,DBLP:conf/aaai/BaiBTL17}.An exhaustive review of previous work is beyond the ability of this paper. Here, we focus on two main streams of deep metric learning, contrastive embedding and triplet embedding, and their recent variants used in computer vision.

 Contrastive embedding requires that training data integrates precise pair-wise similarities or distances, but the requirements are always hard to reach in real practice. To alleviate the problem, triplet embedding Hinton et al.  \cite{DBLP:journals/jmlr/SalakhutdinovH07} is proposed to guide the embedding process with relative similarity information from different pairs and it has been widely used in many tasks, such as image retrieval \cite{DBLP:conf/cvpr/WangSLRWPCW14,DBLP:journals/jmlr/ChechikSSB10}.
 
\subsection{Fine-grained image retrieval}
Fine-grained image retrieval, is a challenging task since we have to deal with fine-grained information of the images, and sometimes the candidates could be so similar that even human could not clearly recognize. Deep metric learning, which learns a distance consistent with a notion of semantic similarity, is powerful to deal with this task and it is also a common application of deep metric learning.

Movshovitz-Attias et al.\cite{DBLP:conf/iccv/Movshovitz-Attias17} use triplet-based methods, but these methods are hard to optimize.  A main problem is the need for finding informative triplets. Movshovitz-Attias et al.\cite{DBLP:conf/iccv/Movshovitz-Attias17} proposed a proxy-based method, which consists of an anchor data point and similar and dissimilar proxy points which are learned as well. Han et al.\cite{DBLP:conf/mm/HanGZZ18} proposed a novel attention model to derive the local information of the image, and make global and local information benefit from each other. Lin et al.\cite{DBLP:conf/eccv/LinDDLZ18} proposed a deep variational metric learning framework to explicitly model the intra-class variance an disentangle the intra-class invariance. By this way, it could simultaneously generate discriminative samples to improve robustness.

Song et al.\cite{DBLP:conf/cvpr/SongJR017} proposed a new metric learning scheme, based on structured prediction, that is aware of the global structure of the embedding space, and which is designed to optimize a clustering quality metric. Xuan et al.\cite{DBLP:conf/eccv/XuanSP18} proposed a method to define a family of embedding functions that can be used as an ensemble to give improved results. Each embedding function is learned by randomly bagging the training labels into small subsets. Yu et al.\cite{DBLP:conf/eccv/YuLGDT18} pointed that the training process of feature embedding with triplet loss is usually very sensitive to the selected triplets, and  randomly selected triplets and selecting hardest triplets also leads to bad local minima. Yu et al.\cite{DBLP:conf/eccv/YuLGDT18} argued that the bias in sampling of triplets degrades the performance of learning with triplet loss. They alleviated the problem by proposing a new variant of triplet loss, which tried to reduce the bias in triplet sampling by adaptively correcting the distribution shift on sampled triplets. 


The works mentioned above mainly focus on how to more robustly and effectively use the visual information of the training data or deal with the inherent problems of deep metric learning loss such as triplet loss, but hardly mines the semantic relationships of images. 
\subsection{Semantic-based image retrieval}
Here, we want to show some existed works which try to introduce some semantic or hierarchical information into the retrieval task, so sometimes the task is not only conducted in a fine-grained way but also get category-level or semantic-level retrieval results.

\cite{DBLP:conf/cvpr/LiuLQWT16,DBLP:journals/corr/abs-1807-11674,DBLP:journals/corr/abs-1901-07973} work on clothes-related tasks such as retrieval and segmentation. They proposed a clothes dataset with multiple labels such as style, scale, viewpoint, occlusion. But from our point, these attributes are actually arranged in a parallel manner. These is no explicit hierarchical relationship between these attributes, so the semantic embedding approaches could not be used on this dataset. Chen et al.\cite{DBLP:conf/mm/ChenWGDLL18} investigated simultaneously predicting categories of different levels in the hierarchy and integrating this structured correlation information into the deep neural network to perform semantic embedding and could benefit fine-grained retrieval. They also collected a four-level category dataset with hierarchical information with butterfly images. 

Our proposed dataset are collected from real application of online image shopping which is beneficial for real practice of novel machine learning algorithms. Besides, some work use the hierarchical information by using the visual distance; Ge et al.(HTL)\cite{DBLP:conf/eccv/GeHDS18} constructed a hierarchical class-level tree where neighboring classes are merged recursively, and they computed the distance to evaluate neighboring classes by the euclidean distance on the feature space, and use this hierarchical tree to select the margins in the triplet loss, but ignoring the semantic information form the annotated hierarchical labels. We get the idea to build the semantic-visual tree based on this work, and we adopt it as our baseline. Besides, our proposed sampling strategy is a variant of the anchor neighbor sampling in \cite{DBLP:conf/eccv/GeHDS18}, we modify it to fit better with our semantic-visual tree.
\subsection{Merchandise image retrieval} 
There has been increasing interests in developing merchandise  image retrieval systems in both research and industry. He et al.\cite{DBLP:conf/cvpr/HeFLCLCC12} proposed a visual search system. Zhan et al.\cite{DBLP:conf/bmvc/ZhanSK17} handle the large visual discrepancy for the same shoe in different forms of online merchandise  image by using multi-task and multi-view network structure and a new loss function to minimize the distances between images of the same shoe from different viewpoints. Since the street image or the user image are always from different viewpoints, how to deal with this gap seems to be important for merchandise  image retrieval. Zhan et al.\cite{DBLP:conf/icip/ZhanSK17} tries to learn the viewpoint-invariant triplet network by training the triplet network at different scales so that the representation of image incorporates different levels of invariance at different scales. 

Another challenging property of merchandise  image retrieval is that, as \cite{DBLP:conf/iccv/HuangFCY15} pointed, the large discrepancy between online shopping images, usually taken in ideal pose/background/lighting conditions, and user images captured in uncontrolled conditions. This paper used a dual structure to deal with the gap.

\section{Method}
\begin{figure}[t]

\centering
    \includegraphics[width=1\linewidth]{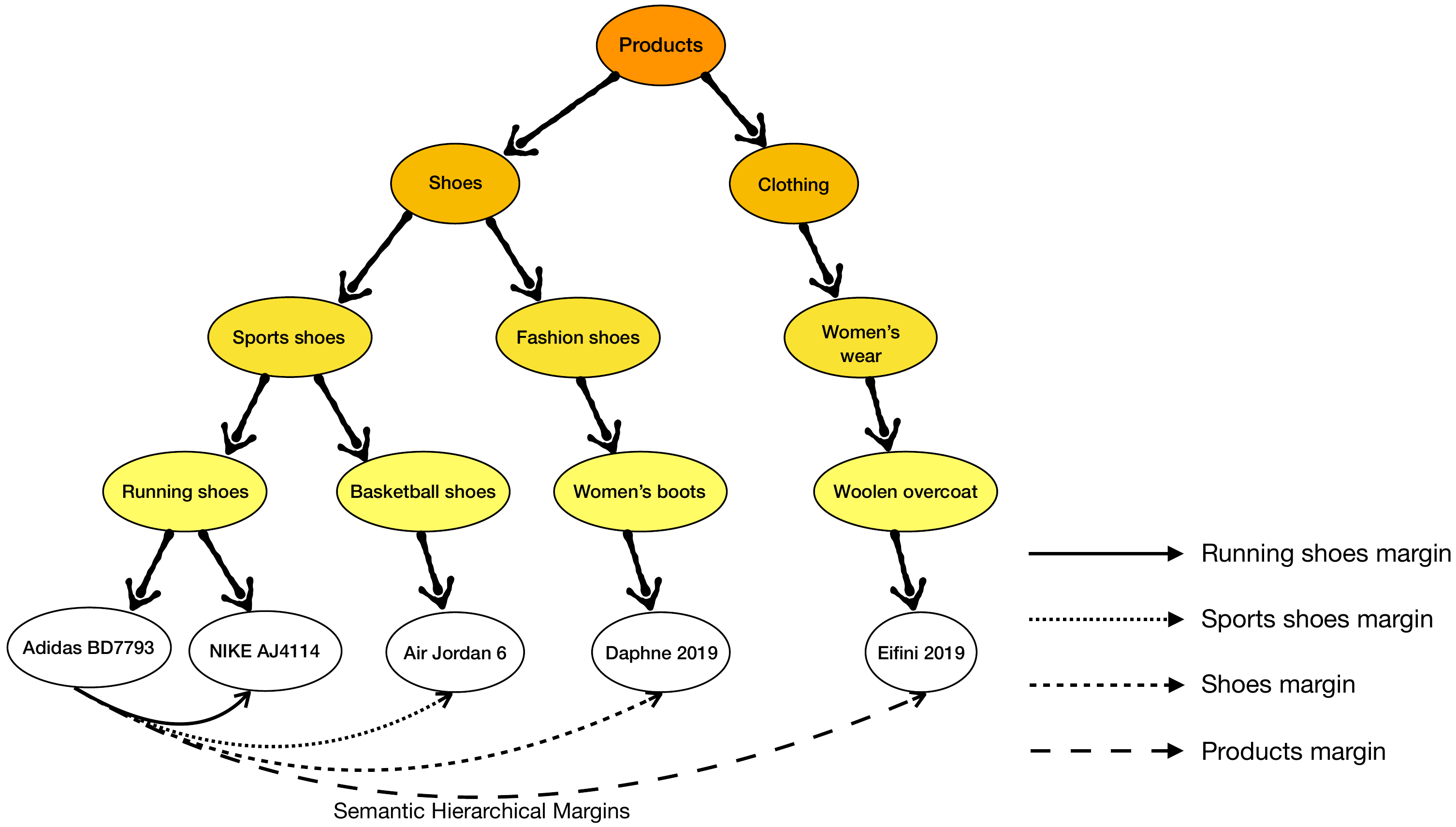}
\caption{This is a toy example of our hierarchical semantic tree. Semantically further classes would have bigger margin in the loss.}
\label{fig_tree}
\end{figure}
We propose a novel deep metric learning framework called hierarchical adaptive semantic-visual tree. The overall structure of our proposed model can be divided into two parts--semantic hierarchy and visual similarity ranking; Semantic hierarchy integrates the multi-level abstraction information of the given labels to modify the margin in the contrastive loss and visual similarity ranking module incorporates the visual information to guide the margin under the control of semantic hierarchy. 

To constrain the visual similarity under semantic classes, a scalar $\alpha$ is added to visual similarity margin $S_{i,j}$, where $i,j$ refer to the the $i^{th}$ and $j^{th}$ category-level semantic class having same parents. The two categories are the children nodes of a third-level node in our ASVT. We will clearly show how to choose the margin of contrastive loss based on our model.
Here are some notations that will be used to describe our model:
\begin{itemize}
\item[*] $\mathcal{P}=\left\{I_i^+,I_j^+\right\}$:all the positive image pairs constructed from training set, where $I_i^+$ and $I_j^+$ are supposed to have the same labels of all levels (fine-grained).
\item[*] $\mathcal{N}=\left\{I_i^-,I_j^-\right\}$:all the negative image pairs constructed from training set.
\item[*] $\mathcal{M}_{i,j}$ the margin between the  $\left\{I_i^-,I_j^-\right\}$.
\item[*] $m$ the semantic margin computed by LCS similarity method.
\item[*] $\mathcal{S}_{i,j}$ visual similarity between the class $i$ and class $j$.
\item[*] $\mathcal{D}\left(I_i,I_j\right)$ Euclidean distance between $I_i$ and $I_j$.
\end{itemize}
\subsection{Contrastive loss}
The goal of the deep metric learning loss encourages semantically similar features to have similar features. And in our paper, we take contrastive loss as feature embedding loss. Contrastive loss is fed with positive and negative pairs, where we want to make $\mathcal{D}\left(I_i^-,I_j^-\right)$ and $\mathcal{D}\left(I_i^+,I_j^+\right)$ smaller s.t. $\mathcal{D}\left(I_i^-,I_j^-\right)\leq  \mathcal{M}$. This loss penalizes large distance between positive pairs and negative pairs with distance smaller than a margin. An appropriate margin parameter is critical to retrieval performance. A small margin cannot integrate sufficient training information, while a large margin may introduce noise that result in over-fitting. That's the reason why margin is critical in the feature embedding task.

A certain setting of margin represents a certain prior perspective of the designer and the comprehension about the data distribution. A simple but not trivial setting is that dissimilar user image - online image pair should have larger margin while the ones only differ in subtle parts need smaller margin. However, for most of the deep metric embedding algorithms, the margin is fixed; the semantic distance for the pairs are thus fixed. When it comes to our proposed JDProduct dataset, using a constant margin for every instance regardless of the semantic hierarchy obviously seems not to be a good choice. And we would introduce in detail how our margin is given according to the context information. Here we give the conventional contrastive loss.

\begin{equation}
\begin{split}
\mathcal{L}=\sum\limits_{(i,j)\in \mathcal{P}}\mathcal{D}\left(f_{i}^+,f_{j}^+\right)+
\sum\limits_{(i,j)\in \mathcal{N}}max\left\{0,\mathcal{M}-\mathcal{D}\left(f_{i}^-,f_{j}^-\right)\right\}
\end{split}
\end{equation}
Where $\mathcal{M}$ is a fixed constant here.
\subsection{Semantic hierarchical loss}
Classes in higher level of the semantic tree should have larger margin due to its semantic dissimilarity. The measure to help us evaluate the dissimilarity should satisfy $m^2>m^3>m^4$. We thus choose a measure based on the tree structure. Suppose $\mathcal{G}=(V,E)$ to be a directed acyclic graph (DAG) with nodes $V$ and edges $E\subset V\times V$. Since it is a directed graph, an edge $(u,v)\in E$ means $v$ is a sub-class of $u$ semantically.

We use a common measure in this DAG structure for the dissimilarity $d_\mathcal{G}$ of classes; we compute the height of the \textit{least common subsumer (LCS)}, divided by the height of the hierarchy:
\begin{equation}
    d_\mathcal{G}=\frac{height(lcs(u,v))}{height(\mathcal{G})}
\end{equation}
where the height of a node is defined as the length of the longest path from that node to a leaf. And the height of a tree is $max_{v\in V}height(v)$. The LCS of two nodes ($lcs(u,v)$) is the ancestor of both nodes $u$ and $v$ that no child is also the ancestor of those nodes (least property).The range of $d_\mathcal{G}$ is $(0,1]$.

 Two classes with higher $d_\mathcal{G}$ should have higher semantic margin. To regulate the dissimilarity to satisfy the strict monotonicity of the margins along with hierarchy level, we also apply a scalar $\gamma$ and a bias $\beta$ to $s_\mathcal{G}$ to form our final semantic hierarchical margin.
\begin{equation}
    m=\gamma\cdot d_\mathcal{G} +\beta
\end{equation}
where $\gamma$ and $\beta$ are hyperparameter that are given initially. 

\begin{figure}[t]
\centering
    \includegraphics[width=1\linewidth]{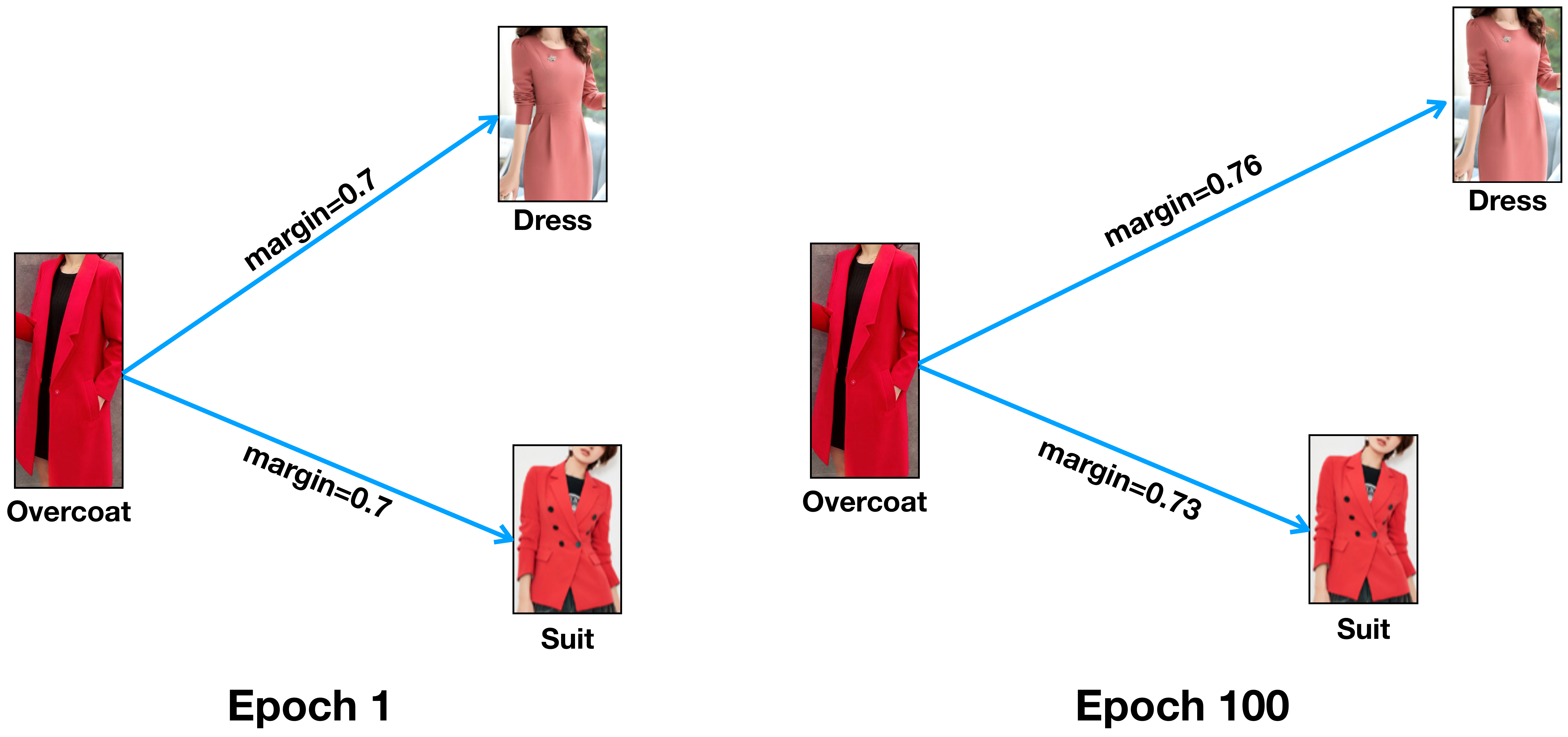}
\caption{The adaptive modifications with visual similarity.Even dress and suit are both in different semantic class with overcoat, but they should not have a same margin with overcoat because of their visual features.}
\label{fig_visualupdate}
\end{figure}

\subsection{Visual similarity ranking }
Visual similarity is crucial for the user-image and online merchandise  matching, especially for the fine-grained retrieval. We assume that users want the merchandise  which is not only functionally similar but also visually similar. Since only using the semantic information may cause visual dissimilarity, this module acts as a constrain under the semantic margins. 

Taking the running shoe as an example, there are hundreds of running shoes which belong to the same fine-grained category but differ pretty much in visually obvious properties such as color and shape. The capability of finding the exact instance is essential, but we still have to list  related results in the following lines, considering the results which may best fit the demand of the user. From this motivation, we add visual similarity rank to our model, not only resulting in better retrieval performance, but also perfectly satisfying our need. 

The benefit of visual similarity rank is shown on Fig.\ref{fig_main} .Without it the results are only semantically related but differ greatly visually. Along with the training process, the feature representations of images would change and we could get different visual similarity between images as in Fig.\ref{fig_visualupdate};

We only compute visual similarity in the category-level, since the higher level differ too much semantically which makes the visual information not informative enough to guide. That is to say, only the margin between class in the category-level will be affected by visual similarity, while the margins of higher levels are fixed. The calculation of visual similarity is given as below:
\begin{equation}
\begin{split}
S_{i,j}=\frac{1}{mn}\times\sum\limits_{\forall x_i,x_j\in \mathcal{N}}||f(x_i)-f(x_j)||_2
\end{split}
\end{equation}
where m is the number of samples in class i and n is the number of samples in class j. $f(x_i),f(x_j)$ are the features extracted by the DNNs with $L_2$ normalization, and the dimension of feature is $d$. $\mathcal{N}$ is mentioned above. Our final margin is the combination of semantic hierarchy and visual similarity ranking as below:
\begin{equation}
\mathcal{M}_{i,j}=m+\alpha\cdot S_{i,j}
\end{equation}
where $\mathcal{M}_{i,j}$ is the final margin of class $i$ and class $j$ (class $i$ and class $j$ have a mutual ancestor in a certain level according to the actual semantic relationship), m is the fixed semantic margin computed form the LCS similarity method. Besides, to constrain visual similarity from interfering the semantic gap, $\alpha$ should be small enough. And here is the our final contrastive loss with ASVT:
\begin{equation}
\begin{split}
\mathcal{L}=\sum\limits_{(i,j)\in \mathcal{P}}\mathcal{D}\left(f_{i}^+,f_{j}^+\right)+
\sum\limits_{(i,j)\in \mathcal{N}}max\left\{0,\mathcal{M}_{i,j}-\mathcal{D}\left(f_{i}^-,f_{j}^-\right)\right\}
\end{split}
\end{equation}

\begin{algorithm}  
        \caption{Training with adaptive semantic-visual tree}  
        \begin{algorithmic} 
            \Require Training data with hierarchical labels $\mathcal{D}={(x_i,(y_i^1,y_i^2...))}$. Network is initialized with a pretrained model on ImageNet. The semantic tree based on dataset and the initialized visual similarity ranking parameters.
            \While{not converge} \do
                \State $epoch \gets epoch+1$  
                \State Use our batch sampling strategy to form our data mini-batch
                \State //Compute $m_{i,j}$ between training sample $i$ and $j$ with LCS method
                \State $d_\mathcal{G} \gets \frac{height(lcs(i,j))}{height(\mathcal{G})}$
                \State $s_\mathcal{G}(i,j) \gets 1-d_\mathcal{G}(i,j)$
                \State $m_{i,j} \gets \gamma\cdot s_\mathcal{G} +\beta$
                \State //Initialize visual similarity with $0$
                \State $S_{i,j} \gets 0$
                \If{epoch>1}
                    \State //Compute $S_{i,j}$ between training sample $i$ and $j$
                    \State $S_{i,j} \gets \frac{1}{mn}\times\sum\limits_{\forall x_i,x_j\in\mathcal{N}}||f(x_i)-f(x_j)||_2$
                    \State //Update the margin $m_{i,j}$ with their visual similarity $S_{i,j}$
                    \State $\mathcal{M}_{i,j} \gets m_{i,j}+\alpha\cdot S_{i,j}$
                \EndIf
                \State Compute loss for sample $i$ and $j$ with $\mathcal{M}_{i,j}$

                \State Backprop the gradients and update learnable parameters.
                \EndWhile
        \end{algorithmic}  
    \end{algorithm} 
    
\subsection{Batch sample}
For the semantic-visual tree which has $S'$ levels, we randomly select one node at each level so we get $S'$ nodes(the parent level of the leaves, to ensure that all the semantic margins appear in a mini-batch). Each node represents an user query image of a certain semantic class, so we can thus maintain the diversity of training samples in a mini-batch, which is so important for training the network with deep metric learning loss. 

Then, to also preserve the visual diversity into a mini-batch, we firstly rank the classes by similarity and choose the top $M'$ classes; here we already get $S'\times M'$ nodes. We finally randomly select $t'$ merchandise  image samples from each class of the $S'\times M'$ nodes, to form our final mini-batch, and the size is $S'\times M'\times t'$. The goal of collecting visually nearest classes is to encourage the model to learn discriminative features from the visually similar classes.

\begin{table}
  \centering
  \caption{Comparative results of different methods on JDProduct dataset.}
  \label{tab:JDProduct}

  \begin{tabular}{lllllll}
    \toprule
    Methods&R@1&R@2&R@4&R@8&R@16&R@32\\
    \midrule
HDC\cite{DBLP:conf/iccv/YuanYZ17}&26.14&56.36&73.03&79.61&82.70&85.90\\
    BIER\cite{DBLP:journals/corr/abs-1801-04815}&29.71&60.34&77.60&83.99&86.04&88.47\\
    HTL\cite{DBLP:conf/eccv/GeHDS18}&31.19&61.65&78.23&85.64&88.58&90.01\\
    our baseline&30.81&61.48&77.11&83.69&87.53&89.55\\
    SH&34.71&66.46&81.01&88.49&91.28&92.64\\
    our model&\textbf{35.04}&\textbf{67.91}&\textbf{83.25}&\textbf{89.17}&\textbf{92.85}&\textbf{94.31}\\
  \bottomrule
\end{tabular}
\end{table}

\begin{table}\small
    \centering
  \caption{Comparison between different hierarchical
levels of the proposed ASVT model.}
  \label{tab:level}
  \begin{tabular}{lllllll}
    \toprule
Hierarchical Level&R@1&R@2&R@4&R@8&R@16&R@32\\
    \midrule
 Fifth Level(Instance)&35.04&67.91&83.25&89.17&92.85&94.31\\
    Fourth Level (Category) &42.66&80.49&92.15&96.25&98.31&99.25\\
    Third Level&47.46&86.73&97.15&98.05&99.17&99.39\\
    Second Level&54.38&92.47&98.86&99.98&1.0&1.0\\
  \bottomrule
\end{tabular}
\end{table}

\begin{figure*}[t]

\centering
    \includegraphics[width=0.8\linewidth]{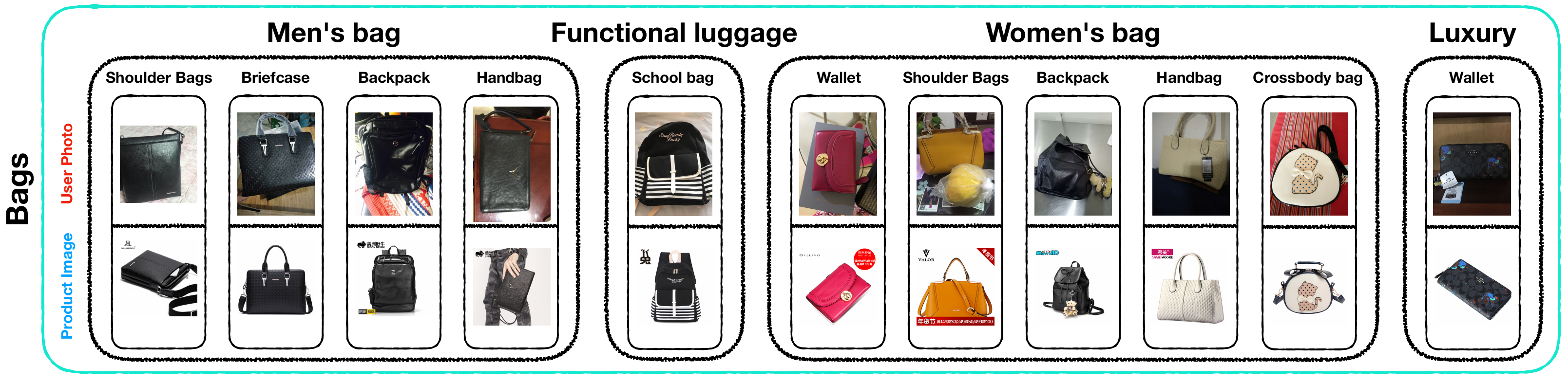}
\caption{A part of our proposed dataset, JDProduct which comes from the actual application of JD.com online shopping platform. We collect all the merchandise from the root node "Outfits", containing 3 second-level categories(Bags,Clothes,shoes), 14 third-level categories, 92 fourth-level categories, and 11990 instance pairs, which include a user query image and a corresponding official merchandise  image. Fig.\ref{fig_tree} is a small part of the semantic tree of JDProduct.}
\label{fig_dataset}
\end{figure*}

\section{JDProduct Dataset}
We contribute JDProduct, a large-scale user image and online merchandise  retrieval dataset, to the community. JDProduct has several appealing properties. First, JDProduct is annotated with four hierarchical labels representing different levels of semantic abstraction.  Second, it also contains a large-scale cross-domain image pairs, each instance-level image is paired with an exact user query image. Some example images along with the annotations are shown in Fig\ref{fig_dataset}. 
\subsection{Image collection}
We collect our dataset based on the image shopping function in JD.com shopping platform. The users of the application would upload actual query image to search for merchandise, and the image is then restored to constitute the user query part of our dataset, and the according merchandise  image makes up another part.
\subsection{Hierarchical annotations}
All the data is collected among four semantic structure, which constitute a merchandise semantic structural tree of height 5 Fig.\ref{fig_tree}.

The first layer of the tree is the root node, the nodes in the second layer are the main types of merchandise divided into three major categories: bags, clothed, and shoes. The nodes in third layer are finer divisions of merchandise, in total, we have 14 categories in this level (e.g., Men's, Women's clothing, Sportswear, Children's wear...). The fourth level is category level for merchandise; And the fifth level is instance level where we have lots instances of their corresponding parents node.

Totally we have 11990 instances in our dataset. The nodes closer to the root of the semantic tree refer to more abstract concepts while nodes closer to the leaves indicate fine-grained concepts. 

\section{Experiments}

\subsection{Implementation}
All of our experiments are implemented with Caffe framework. The backbone network is GoogLeNet with batch normalization  pre-trained on the ImageNet dataset. The 1000-way fully connected layer is replaced by a 512 dimensional fully connected layer. The new added layer is initialized using the "Xaiver" initializer. The input images are resized and cropped into 224 - 224, and then subtract the mean value. We use the standard SGD as our optimizer.

\subsection{Experiments on JDProduct}
Our proposed JDProduct dataset has 11990 user image - merchandise  image pairs of data, we divide the data as 6990 pairs for training and 5000 pairs for testing. \\
\textbf{Baseline}. 
First, we introduce the baseline we use for our proposed JDProduct dataset. The backbone network of our baseline is SENet\cite{DBLP:conf/cvpr/HuSS18} with standard contrastive loss.
We apply three deep metric learning methods, HDC\cite{DBLP:conf/iccv/YuanYZ17}, BIER\cite{DBLP:journals/corr/abs-1801-04815}, and HTL\cite{DBLP:conf/eccv/GeHDS18} on our JDProduct model as comparative results. These three methods all adopted GoogLeNet as basenet.\\
\textbf{Metric}.
The R@k metric we use for CUB-200-2011 and CARS196 is that we calculate the percentage
of the testing examples whose k nearest neighbors contain at least one example of the same class. And for JDProduct, we calculate the percentage
of the testing examples whose k nearest neighbors contain at least one example of the same instance.
For the experiments on different hierarchical level, we calculate that of the instance, category, third-level label, second-level label, from the bottom to top.
Here we have R@1, R@2, R@4, R@8, R@16, R@32, to evaluate the effectiveness of different methods on JDProduct dataset, and demonstrate the superiority of our proposed model. 

We also did ablation study, where only the semantic hierarchy is used without visual similarity to fully demonstrate the effectiveness of our proposed model. 

Since our JDProduct dataset came from real application, the user images are often from multi-viewpoints and the background,lighting and pose varied much. From Table.\ref{tab:JDProduct} HDC\cite{DBLP:conf/iccv/YuanYZ17} only got 26.14\% on R@1, which indicates the difficulty of training an effective model on our dataset of real application.

HTL\cite{DBLP:conf/eccv/GeHDS18} performed the best in Table.\ref{tab:JDProduct} among the three previous methods, which may result from the visual hierarchy structure it used that inherently fit the semantic structure of the categories. HTL got around 5\% improvement than HDC. They all used GoogLeNet with similar experiment setting and training procedure, so it is reasonable to imply that methods with hierarchical structures could perform better on our dataset. 

We could see from the ablation study in Table.\ref{tab:JDProduct} (where SH represents semantic hierarchy model) that our semantic hierarchy model(SH)  could improve the baseline by around 4\%, and with visual information the performance could increase even more. But, what visual similarity brings most is not shown on the accuracy, but from visual results as in Fig.\ref{fig_main}.

From Table.\ref{tab:level}, we conducted experiments on different hierarchical level of our ASVT model to give the base accuracy of our JDProduct dataset. The $k^{th}$ level denotes that the retrieval result has the same categories as the query from $k^{th}$ level, as our dataset provides four levels of labels, with the root, we get a tree of height 5. So the fifth level is the instance level. We could see from the results that the accuracy increase rapidly along with the level increasing, which implies that it is easier to learn the high-level abstract information than the fine-grained information. And even with our novel model we could only get 35.04\% the R@1 accuracy of instance-level, which shows that there could be plenty of work to do on our proposed dataset, including multi-viewpoints adaptation, cross domain learning, or adding a prior comprehension of merchandise  image retrieval to leverage the improvement and development of modern machine learning algorithms and enhance the application in industrial world.

\begin{figure*}[t]

\centering
    \includegraphics[width=0.8\linewidth]{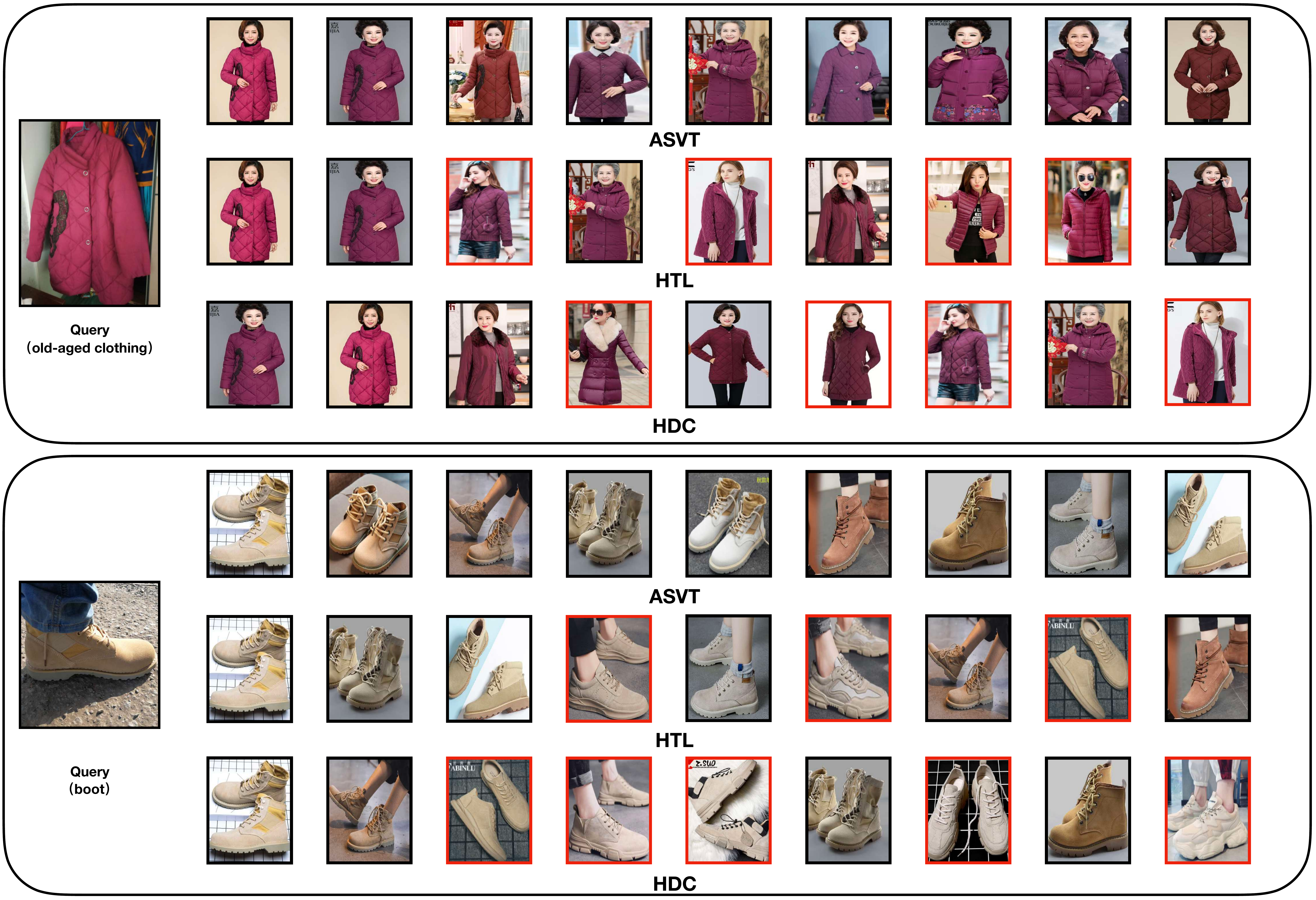}
\caption{The comparative retrieval results on JDProduct dataset. The first row is the results of our ASVT model,  the second row is the results of HTL and the third row is HDC's result. The semantic mismatched results are in red frame(old-aged clothing vs young ladies' clothing and high boot vs low-top casual shoes). } 
\label{fig_JDresult}
\end{figure*}

We also show the retrieval results comparing our ASVT model with HTL(only visual similarity) as in Fig.\ref{fig_JDresult}. We could see from the figure that with only visual similarity as the HTL (second row), the retrieval results may contain images from other semantic classes since they are visually more similar. But our ASVT model alleviates this problem and make semantic similarity prior to visual similarity. We could see from the second row, using "old-aged clothing" as query, with HTL there are many "young-aged clothing" as in red frame while with our ASVT model we could only see "old-aged clothing", which demonstrates the effectiveness of our semantic hierarchical loss.
\begin{table}
  \caption{The comparative results on CARS196 dataset.}
  \label{tab:car}
  \begin{tabular}{lllllll}
    \toprule
    Methods&R@1&R@2&R@4&R@8&R@16&R@32\\
    \midrule
    LiftedStruct\cite{DBLP:conf/cvp/SongXJS16}&49.0&60.3&72.1&81.5&89.2&92.8\\
    HDC\cite{DBLP:conf/iccv/YuanYZ17}&73.7&83.2&89.5&93.8&96.7&98.4\\
    BIER\cite{DBLP:journals/corr/abs-1801-04815}&78.0&85.8&91.1&95.1&97.3&98.7\\
    HTL\cite{DBLP:conf/eccv/GeHDS18}&81.4&88.0&92.7&95.7&97.4&99.0\\
  
    Angular loss\cite{DBLP:conf/iccv/WangZWLL17}&71.4&81.4&87.5&92.1&-&-\\
    
    our model&\textbf{86.5}&\textbf{93.0}&\textbf{95.7}&\textbf{97.3}&\textbf{98.9}&\textbf{99.8}\\
  \bottomrule
\end{tabular}
\end{table}

\begin{table}\small
  \caption{Comparison with state-of-art methods on CUB-200-2011 dataset.}
  \label{tab:cub}
  \begin{tabular}{lllllll}
    \toprule
    Instance-Level Recall&R@1&R@2&R@4&R@8&R@16&R@32\\
    \midrule
 LiftedStruct\cite{DBLP:conf/cvp/SongXJS16}&47.2&58.9&70.2&80.2&89.3&93.2\\
Binomial Deviance\cite{DBLP:conf/nips/UstinovaL16}&52.8&64.4&74.7&83.9&90.4&94.3\\
Histogram Loss\cite{DBLP:conf/nips/UstinovaL16}&50.3&61.9&72.6&82.4&88.8&93.7\\
HDC\cite{DBLP:conf/iccv/YuanYZ17}&53.6&65.7&77.0&85.6&91.5&95.5\\
BIER\cite{DBLP:journals/corr/abs-1801-04815}&55.3&67.2&76.9&85.1&91.7&95.5\\
HTL\cite{DBLP:conf/eccv/GeHDS18}&57.1&68.8&78.7&86.5&92.5&95.5\\
Angular loss\cite{DBLP:conf/iccv/WangZWLL17}&54.7&66.3&76.0&83.9&-&-\\

our model&\textbf{60.3}&\textbf{70.1}&\textbf{80.6}&\textbf{89.3}&\textbf{93.7}&\textbf{97.1}\\

  \bottomrule
\end{tabular}
\end{table}

\subsection{Experiments on retrieval datasets}
Retrieval datasets we use:
\begin{itemize}
    \item \textit{CUB-200-2011.} CUB-200-2011 dataset has 200 species of birds with 11,788 images, where the first 100 classes
are for training (5,864 images) and the rest of classes
are for testing (5,924 images). Both query set and database set are the test set
\item \textit{CARS196.} CARS196 dataset has 196 classes of cars
with 16,185 images, where the first 98 classes are for
training (8,054 images) and the other 98 classes are for
testing (8,131 images). Both query set and database set
are the test set.
\end{itemize}
From the comparative results of Table.\ref{tab:car} and Table.\ref{tab:cub} , we could see that our model surpasses the state-of-the-art approaches. Considering the R@1 accuracy, our model has a 5\% improvement in CARS196 dataset than HTL, and 3\% in CUB-200-2011 which shows that out semantic-visual tree has a great advantage over merely using visual hierarchical information. Only with visual information may cause the problem that the information from images could be misleading if two objects are similar but actually different entities or categories. 

For example, SUV cars have similar shapes and sizes but if they are from different brands they are actually totally different instances, if we only guide our model with visual similarity, the network may retrieve the SUV car from different brands, which is a big problem. But our semantic hierarchy based model could alleviate this problem by choosing different margins, which could make the semantically distant images far from each other, reducing the possibility of making such mistake\ref{fig_main}. 

What the angular loss\cite{DBLP:conf/iccv/WangZWLL17} does is that it proposed a structure to form stable triangle during learning process with modifications to the gradients; Compared with angular loss, we have around 5\% improvement in CUB-200-2011 but a remarkable 15\% increase in CARS196. We all use the GoogLeNet, and deep metric learning loss, but by simply setting the margin according to semantic distance there could be a huge increase without introducing extra supervision information.


Since as mentioned above, CARS196 dataset already had label in the form of "BMW-X6-SUV", all we have to do is to do some string manipulation. So we have a parent node "BMW" and a grandparents' node "SUV" of the child node "BMW" which make up a three-level hierarchical structure(The same goes for CUB-200-2011). And from another perspective, our semantic hierarchy is actually a full usage of the existed supervision information that is already embedded in the training data but nobody noticed before. And we provide a novel way to use this inherent hierarchical information without extra-dictionary or a prior information, which may boost the development of all kinds of feature embedding task such as recognition or retrieval. 

The success of our proposed model is not something incredible, but with this perspective of mining the inherent supervision embedded in the labels of training data, the modern deep metric learning techniques could simply add our module to their margin selection procedure and got a remarkable improvement in the results, which is a great contribution to the community. What is more important is that, we could see merely using the dataset such as CARS196 or CUB-200-2011 we could only get a hierarchy of two levels and this hinders the application and development of semantic feature embedding, so JDProduct dataset we provided with five levels could alleviate this problem and to encourage more insights and research to come into this area.
\subsection{Conclusion}
In this paper we collected a user image and merchandise  image retrieval dataset called JDProduct, which comes from real application of image shopping. All the data are given in pairs of a user query and an merchandise  image accordingly. 

Based on our dataset we also proposed a novel adaptive visual-semantic strategy to learn a good feature embedding in the task of fine-grained image retrieval and fine-grained merchandise  image retrieval. In our model, we first build a hierarchical semantic tree based on the hierarchical 
labels and compute the semantic distance by using LCS method, and compute the visual distance adaptively for each epoch to construct the final margin between semantic classes of each level. The superiority of our method over existing state-of-the-art work is
verified on several benchmark datasets.


%
\bibliographystyle{ACM-Reference-Format}
\balance
\bibliography{main}


\begin{thebibliography}{31}


\ifx \showCODEN    \undefined \def \showCODEN     #1{\unskip}     \fi
\ifx \showDOI      \undefined \def \showDOI       #1{#1}\fi
\ifx \showISBNx    \undefined \def \showISBNx     #1{\unskip}     \fi
\ifx \showISBNxiii \undefined \def \showISBNxiii  #1{\unskip}     \fi
\ifx \showISSN     \undefined \def \showISSN      #1{\unskip}     \fi
\ifx \showLCCN     \undefined \def \showLCCN      #1{\unskip}     \fi
\ifx \shownote     \undefined \def \shownote      #1{#1}          \fi
\ifx \showarticletitle \undefined \def \showarticletitle #1{#1}   \fi
\ifx \showURL      \undefined \def \showURL       {\relax}        \fi
\providecommand\bibfield[2]{#2}
\providecommand\bibinfo[2]{#2}
\providecommand\natexlab[1]{#1}
\providecommand\showeprint[2][]{arXiv:#2}

\bibitem[\protect\citeauthoryear{Bai, Bai, Tian, and Latecki}{Bai
  et~al\mbox{.}}{2017}]%
        {DBLP:conf/aaai/BaiBTL17}
\bibfield{author}{\bibinfo{person}{Song Bai}, \bibinfo{person}{Xiang Bai},
  \bibinfo{person}{Qi Tian}, {and} \bibinfo{person}{Longin~Jan Latecki}.}
  \bibinfo{year}{2017}\natexlab{}.
\newblock \showarticletitle{Regularized Diffusion Process for Visual
  Retrieval}. In \bibinfo{booktitle}{\emph{{AAAI}}}. \bibinfo{publisher}{{AAAI}
  Press}, \bibinfo{pages}{3967--3973}.
\newblock


\bibitem[\protect\citeauthoryear{Chechik, Sharma, Shalit, and Bengio}{Chechik
  et~al\mbox{.}}{2010}]%
        {DBLP:journals/jmlr/ChechikSSB10}
\bibfield{author}{\bibinfo{person}{Gal Chechik}, \bibinfo{person}{Varun
  Sharma}, \bibinfo{person}{Uri Shalit}, {and} \bibinfo{person}{Samy Bengio}.}
  \bibinfo{year}{2010}\natexlab{}.
\newblock \showarticletitle{Large Scale Online Learning of Image Similarity
  Through Ranking}.
\newblock \bibinfo{journal}{\emph{Journal of Machine Learning Research}}
  \bibinfo{volume}{11} (\bibinfo{year}{2010}), \bibinfo{pages}{1109--1135}.
\newblock


\bibitem[\protect\citeauthoryear{Chen, Ge, Feng, Xu, and Yang}{Chen
  et~al\mbox{.}}{2018a}]%
        {DBLP:journals/access/ChenGFXY18}
\bibfield{author}{\bibinfo{person}{Min Chen}, \bibinfo{person}{Yongxin Ge},
  \bibinfo{person}{Xin Feng}, \bibinfo{person}{Chuanyun Xu}, {and}
  \bibinfo{person}{Dan Yang}.} \bibinfo{year}{2018}\natexlab{a}.
\newblock \showarticletitle{Person Re-Identification by Pose Invariant Deep
  Metric Learning With Improved Triplet Loss}.
\newblock \bibinfo{journal}{\emph{{IEEE} Access}}  \bibinfo{volume}{6}
  (\bibinfo{year}{2018}), \bibinfo{pages}{68089--68095}.
\newblock


\bibitem[\protect\citeauthoryear{Chen, Wu, Gao, Dong, Luo, and Lin}{Chen
  et~al\mbox{.}}{2018b}]%
        {DBLP:conf/mm/ChenWGDLL18}
\bibfield{author}{\bibinfo{person}{Tianshui Chen}, \bibinfo{person}{Wenxi Wu},
  \bibinfo{person}{Yuefang Gao}, \bibinfo{person}{Le Dong},
  \bibinfo{person}{Xiaonan Luo}, {and} \bibinfo{person}{Liang Lin}.}
  \bibinfo{year}{2018}\natexlab{b}.
\newblock \showarticletitle{Fine-Grained Representation Learning and
  Recognition by Exploiting Hierarchical Semantic Embedding}. In
  \bibinfo{booktitle}{\emph{{ACM} Multimedia}}. \bibinfo{publisher}{{ACM}},
  \bibinfo{pages}{2023--2031}.
\newblock


\bibitem[\protect\citeauthoryear{Chen, Chen, Zhang, and Huang}{Chen
  et~al\mbox{.}}{2017}]%
        {DBLP:conf/cvpr/ChenCZH17}
\bibfield{author}{\bibinfo{person}{Weihua Chen}, \bibinfo{person}{Xiaotang
  Chen}, \bibinfo{person}{Jianguo Zhang}, {and} \bibinfo{person}{Kaiqi Huang}.}
  \bibinfo{year}{2017}\natexlab{}.
\newblock \showarticletitle{Beyond Triplet Loss: {A} Deep Quadruplet Network
  for Person Re-identification}. In \bibinfo{booktitle}{\emph{{CVPR}}}.
  \bibinfo{publisher}{{IEEE} Computer Society}, \bibinfo{pages}{1320--1329}.
\newblock


\bibitem[\protect\citeauthoryear{Dai, Zhang, Wang, Lu, and Wang}{Dai
  et~al\mbox{.}}{2019}]%
        {DBLP:journals/tip/DaiZWLW19}
\bibfield{author}{\bibinfo{person}{Ju Dai}, \bibinfo{person}{Pingping Zhang},
  \bibinfo{person}{Dong Wang}, \bibinfo{person}{Huchuan Lu}, {and}
  \bibinfo{person}{Hongyu Wang}.} \bibinfo{year}{2019}\natexlab{}.
\newblock \showarticletitle{Video Person Re-Identification by Temporal Residual
  Learning}.
\newblock \bibinfo{journal}{\emph{{IEEE} Trans. Image Processing}}
  \bibinfo{volume}{28}, \bibinfo{number}{3} (\bibinfo{year}{2019}),
  \bibinfo{pages}{1366--1377}.
\newblock


\bibitem[\protect\citeauthoryear{Ge, Huang, Dong, and Scott}{Ge
  et~al\mbox{.}}{2018}]%
        {DBLP:conf/eccv/GeHDS18}
\bibfield{author}{\bibinfo{person}{Weifeng Ge}, \bibinfo{person}{Weilin Huang},
  \bibinfo{person}{Dengke Dong}, {and} \bibinfo{person}{Matthew~R. Scott}.}
  \bibinfo{year}{2018}\natexlab{}.
\newblock \showarticletitle{Deep Metric Learning with Hierarchical Triplet
  Loss}. In \bibinfo{booktitle}{\emph{{ECCV} {(6)}}}
  \emph{(\bibinfo{series}{Lecture Notes in Computer Science})},
  Vol.~\bibinfo{volume}{11210}. \bibinfo{publisher}{Springer},
  \bibinfo{pages}{272--288}.
\newblock


\bibitem[\protect\citeauthoryear{Ge, Zhang, Wu, Wang, Tang, and Luo}{Ge
  et~al\mbox{.}}{2019}]%
        {DBLP:journals/corr/abs-1901-07973}
\bibfield{author}{\bibinfo{person}{Yuying Ge}, \bibinfo{person}{Ruimao Zhang},
  \bibinfo{person}{Lingyun Wu}, \bibinfo{person}{Xiaogang Wang},
  \bibinfo{person}{Xiaoou Tang}, {and} \bibinfo{person}{Ping Luo}.}
  \bibinfo{year}{2019}\natexlab{}.
\newblock \showarticletitle{DeepFashion2: {A} Versatile Benchmark for
  Detection, Pose Estimation, Segmentation and Re-Identification of Clothing
  Images}.
\newblock \bibinfo{journal}{\emph{CoRR}}  \bibinfo{volume}{abs/1901.07973}
  (\bibinfo{year}{2019}).
\newblock


\bibitem[\protect\citeauthoryear{Han, Guo, Zhang, and Zhu}{Han
  et~al\mbox{.}}{2018}]%
        {DBLP:conf/mm/HanGZZ18}
\bibfield{author}{\bibinfo{person}{Kai Han}, \bibinfo{person}{Jianyuan Guo},
  \bibinfo{person}{Chao Zhang}, {and} \bibinfo{person}{Mingjian Zhu}.}
  \bibinfo{year}{2018}\natexlab{}.
\newblock \showarticletitle{Attribute-Aware Attention Model for Fine-grained
  Representation Learning}. In \bibinfo{booktitle}{\emph{{ACM} Multimedia}}.
  \bibinfo{publisher}{{ACM}}, \bibinfo{pages}{2040--2048}.
\newblock


\bibitem[\protect\citeauthoryear{He, Feng, Liu, Cheng, Lin, Chung, and
  Chang}{He et~al\mbox{.}}{2012}]%
        {DBLP:conf/cvpr/HeFLCLCC12}
\bibfield{author}{\bibinfo{person}{Junfeng He}, \bibinfo{person}{Jinyuan Feng},
  \bibinfo{person}{Xianglong Liu}, \bibinfo{person}{Tao Cheng},
  \bibinfo{person}{Tai{-}Hsu Lin}, \bibinfo{person}{Hyunjin Chung}, {and}
  \bibinfo{person}{Shih{-}Fu Chang}.} \bibinfo{year}{2012}\natexlab{}.
\newblock \showarticletitle{Mobile product search with Bag of Hash Bits and
  boundary reranking}. In \bibinfo{booktitle}{\emph{{CVPR}}}.
  \bibinfo{publisher}{{IEEE} Computer Society}, \bibinfo{pages}{3005--3012}.
\newblock


\bibitem[\protect\citeauthoryear{Hu, Shen, and Sun}{Hu et~al\mbox{.}}{2018}]%
        {DBLP:conf/cvpr/HuSS18}
\bibfield{author}{\bibinfo{person}{Jie Hu}, \bibinfo{person}{Li Shen}, {and}
  \bibinfo{person}{Gang Sun}.} \bibinfo{year}{2018}\natexlab{}.
\newblock \showarticletitle{Squeeze-and-Excitation Networks}. In
  \bibinfo{booktitle}{\emph{{CVPR}}}. \bibinfo{publisher}{{IEEE} Computer
  Society}, \bibinfo{pages}{7132--7141}.
\newblock


\bibitem[\protect\citeauthoryear{Huang, Feris, Chen, and Yan}{Huang
  et~al\mbox{.}}{2015}]%
        {DBLP:conf/iccv/HuangFCY15}
\bibfield{author}{\bibinfo{person}{Junshi Huang},
  \bibinfo{person}{Rog{\'{e}}rio~Schmidt Feris}, \bibinfo{person}{Qiang Chen},
  {and} \bibinfo{person}{Shuicheng Yan}.} \bibinfo{year}{2015}\natexlab{}.
\newblock \showarticletitle{Cross-Domain Image Retrieval with a Dual
  Attribute-Aware Ranking Network}. In \bibinfo{booktitle}{\emph{{ICCV}}}.
  \bibinfo{publisher}{{IEEE} Computer Society}, \bibinfo{pages}{1062--1070}.
\newblock


\bibitem[\protect\citeauthoryear{Kalenichenko and Philbin}{Kalenichenko and
  Philbin}{2015}]%
        {DBLP:conf/cvpr/SchroffKP15}
\bibfield{author}{\bibinfo{person}{Florian Schroff~Dmitry Kalenichenko} {and}
  \bibinfo{person}{James Philbin}.} \bibinfo{year}{2015}\natexlab{}.
\newblock \showarticletitle{FaceNet: {A} unified embedding for face recognition
  and clustering}. In \bibinfo{booktitle}{\emph{{CVPR}}}.
  \bibinfo{publisher}{{IEEE} Computer Society}, \bibinfo{pages}{815--823}.
\newblock


\bibitem[\protect\citeauthoryear{Lin, Duan, Dong, Lu, and Zhou}{Lin
  et~al\mbox{.}}{2018}]%
        {DBLP:conf/eccv/LinDDLZ18}
\bibfield{author}{\bibinfo{person}{Xudong Lin}, \bibinfo{person}{Yueqi Duan},
  \bibinfo{person}{Qiyuan Dong}, \bibinfo{person}{Jiwen Lu}, {and}
  \bibinfo{person}{Jie Zhou}.} \bibinfo{year}{2018}\natexlab{}.
\newblock \showarticletitle{Deep Variational Metric Learning}. In
  \bibinfo{booktitle}{\emph{{ECCV} {(15)}}} \emph{(\bibinfo{series}{Lecture
  Notes in Computer Science})}, Vol.~\bibinfo{volume}{11219}.
  \bibinfo{publisher}{Springer}, \bibinfo{pages}{714--729}.
\newblock


\bibitem[\protect\citeauthoryear{Liu, Ma, Wang, and Wang}{Liu
  et~al\mbox{.}}{2018}]%
        {DBLP:journals/pr/LiuMWW18}
\bibfield{author}{\bibinfo{person}{Xiaokai Liu}, \bibinfo{person}{Xiaorui Ma},
  \bibinfo{person}{Jie Wang}, {and} \bibinfo{person}{Hongyu Wang}.}
  \bibinfo{year}{2018}\natexlab{}.
\newblock \showarticletitle{M\({}^{\mbox{3}}\)L: Multi-modality mining for
  metric learning in person re-Identification}.
\newblock \bibinfo{journal}{\emph{Pattern Recognition}}  \bibinfo{volume}{76}
  (\bibinfo{year}{2018}), \bibinfo{pages}{650--661}.
\newblock


\bibitem[\protect\citeauthoryear{Liu, Luo, Qiu, Wang, and Tang}{Liu
  et~al\mbox{.}}{2016}]%
        {DBLP:conf/cvpr/LiuLQWT16}
\bibfield{author}{\bibinfo{person}{Ziwei Liu}, \bibinfo{person}{Ping Luo},
  \bibinfo{person}{Shi Qiu}, \bibinfo{person}{Xiaogang Wang}, {and}
  \bibinfo{person}{Xiaoou Tang}.} \bibinfo{year}{2016}\natexlab{}.
\newblock \showarticletitle{DeepFashion: Powering Robust Clothes Recognition
  and Retrieval with Rich Annotations}. In \bibinfo{booktitle}{\emph{{CVPR}}}.
  \bibinfo{publisher}{{IEEE} Computer Society}, \bibinfo{pages}{1096--1104}.
\newblock


\bibitem[\protect\citeauthoryear{Movshovitz{-}Attias, Toshev, Leung, Ioffe, and
  Singh}{Movshovitz{-}Attias et~al\mbox{.}}{2017}]%
        {DBLP:conf/iccv/Movshovitz-Attias17}
\bibfield{author}{\bibinfo{person}{Yair Movshovitz{-}Attias},
  \bibinfo{person}{Alexander Toshev}, \bibinfo{person}{Thomas~K. Leung},
  \bibinfo{person}{Sergey Ioffe}, {and} \bibinfo{person}{Saurabh Singh}.}
  \bibinfo{year}{2017}\natexlab{}.
\newblock \showarticletitle{No Fuss Distance Metric Learning Using Proxies}. In
  \bibinfo{booktitle}{\emph{{ICCV}}}. \bibinfo{publisher}{{IEEE} Computer
  Society}, \bibinfo{pages}{360--368}.
\newblock


\bibitem[\protect\citeauthoryear{Nguyen, Maclagan, Nguyen, Nguyen, Flemons,
  Andrews, Ritchie, and Phung}{Nguyen et~al\mbox{.}}{2017}]%
        {DBLP:conf/dsaa/NguyenMNNFARP17}
\bibfield{author}{\bibinfo{person}{Hung Nguyen}, \bibinfo{person}{Sarah~J.
  Maclagan}, \bibinfo{person}{Tu~Dinh Nguyen}, \bibinfo{person}{Thin Nguyen},
  \bibinfo{person}{Paul Flemons}, \bibinfo{person}{Kylie Andrews},
  \bibinfo{person}{Euan~G. Ritchie}, {and} \bibinfo{person}{Dinh~Q. Phung}.}
  \bibinfo{year}{2017}\natexlab{}.
\newblock \showarticletitle{Animal Recognition and Identification with Deep
  Convolutional Neural Networks for Automated Wildlife Monitoring}. In
  \bibinfo{booktitle}{\emph{{DSAA}}}. \bibinfo{publisher}{{IEEE}},
  \bibinfo{pages}{40--49}.
\newblock


\bibitem[\protect\citeauthoryear{Opitz, Waltner, Possegger, and Bischof}{Opitz
  et~al\mbox{.}}{2018}]%
        {DBLP:journals/corr/abs-1801-04815}
\bibfield{author}{\bibinfo{person}{Michael Opitz}, \bibinfo{person}{Georg
  Waltner}, \bibinfo{person}{Horst Possegger}, {and} \bibinfo{person}{Horst
  Bischof}.} \bibinfo{year}{2018}\natexlab{}.
\newblock \showarticletitle{Deep Metric Learning with {BIER:} Boosting
  Independent Embeddings Robustly}.
\newblock \bibinfo{journal}{\emph{CoRR}}  \bibinfo{volume}{abs/1801.04815}
  (\bibinfo{year}{2018}).
\newblock


\bibitem[\protect\citeauthoryear{Salakhutdinov and Hinton}{Salakhutdinov and
  Hinton}{2007}]%
        {DBLP:journals/jmlr/SalakhutdinovH07}
\bibfield{author}{\bibinfo{person}{Ruslan Salakhutdinov} {and}
  \bibinfo{person}{Geoffrey~E. Hinton}.} \bibinfo{year}{2007}\natexlab{}.
\newblock \showarticletitle{Learning a Nonlinear Embedding by Preserving Class
  Neighbourhood Structure}. In \bibinfo{booktitle}{\emph{{AISTATS}}}
  \emph{(\bibinfo{series}{{JMLR} Proceedings})}, Vol.~\bibinfo{volume}{2}.
  \bibinfo{publisher}{JMLR.org}, \bibinfo{pages}{412--419}.
\newblock


\bibitem[\protect\citeauthoryear{Song, Jegelka, Rathod, and Murphy}{Song
  et~al\mbox{.}}{2017}]%
        {DBLP:conf/cvpr/SongJR017}
\bibfield{author}{\bibinfo{person}{Hyun~Oh Song}, \bibinfo{person}{Stefanie
  Jegelka}, \bibinfo{person}{Vivek Rathod}, {and} \bibinfo{person}{Kevin
  Murphy}.} \bibinfo{year}{2017}\natexlab{}.
\newblock \showarticletitle{Deep Metric Learning via Facility Location}. In
  \bibinfo{booktitle}{\emph{{CVPR}}}. \bibinfo{publisher}{{IEEE} Computer
  Society}, \bibinfo{pages}{2206--2214}.
\newblock


\bibitem[\protect\citeauthoryear{Song, Xiang, Jegelka, and Savarese}{Song
  et~al\mbox{.}}{2016}]%
        {DBLP:conf/cvp/SongXJS16}
\bibfield{author}{\bibinfo{person}{Hyun~Oh Song}, \bibinfo{person}{Yu Xiang},
  \bibinfo{person}{Stefanie Jegelka}, {and} \bibinfo{person}{Silvio Savarese}.}
  \bibinfo{year}{2016}\natexlab{}.
\newblock \showarticletitle{Deep Metric Learning via Lifted Structured Feature
  Embedding}. In \bibinfo{booktitle}{\emph{{CVPR}}}. \bibinfo{publisher}{{IEEE}
  Computer Society}, \bibinfo{pages}{4004--4012}.
\newblock


\bibitem[\protect\citeauthoryear{Ustinova and Lempitsky}{Ustinova and
  Lempitsky}{2016}]%
        {DBLP:conf/nips/UstinovaL16}
\bibfield{author}{\bibinfo{person}{Evgeniya Ustinova} {and}
  \bibinfo{person}{Victor~S. Lempitsky}.} \bibinfo{year}{2016}\natexlab{}.
\newblock \showarticletitle{Learning Deep Embeddings with Histogram Loss}. In
  \bibinfo{booktitle}{\emph{{NIPS}}}. \bibinfo{pages}{4170--4178}.
\newblock


\bibitem[\protect\citeauthoryear{Wang, Song, Leung, Rosenberg, Wang, Philbin,
  Chen, and Wu}{Wang et~al\mbox{.}}{2014}]%
        {DBLP:conf/cvpr/WangSLRWPCW14}
\bibfield{author}{\bibinfo{person}{Jiang Wang}, \bibinfo{person}{Yang Song},
  \bibinfo{person}{Thomas Leung}, \bibinfo{person}{Chuck Rosenberg},
  \bibinfo{person}{Jingbin Wang}, \bibinfo{person}{James Philbin},
  \bibinfo{person}{Bo Chen}, {and} \bibinfo{person}{Ying Wu}.}
  \bibinfo{year}{2014}\natexlab{}.
\newblock \showarticletitle{Learning Fine-Grained Image Similarity with Deep
  Ranking}. In \bibinfo{booktitle}{\emph{{CVPR}}}. \bibinfo{publisher}{{IEEE}
  Computer Society}, \bibinfo{pages}{1386--1393}.
\newblock


\bibitem[\protect\citeauthoryear{Wang, Zhou, Wen, Liu, and Lin}{Wang
  et~al\mbox{.}}{2017}]%
        {DBLP:conf/iccv/WangZWLL17}
\bibfield{author}{\bibinfo{person}{Jian Wang}, \bibinfo{person}{Feng Zhou},
  \bibinfo{person}{Shilei Wen}, \bibinfo{person}{Xiao Liu}, {and}
  \bibinfo{person}{Yuanqing Lin}.} \bibinfo{year}{2017}\natexlab{}.
\newblock \showarticletitle{Deep Metric Learning with Angular Loss}. In
  \bibinfo{booktitle}{\emph{{ICCV}}}. \bibinfo{publisher}{{IEEE} Computer
  Society}, \bibinfo{pages}{2612--2620}.
\newblock


\bibitem[\protect\citeauthoryear{Xuan, Souvenir, and Pless}{Xuan
  et~al\mbox{.}}{2018}]%
        {DBLP:conf/eccv/XuanSP18}
\bibfield{author}{\bibinfo{person}{Hong Xuan}, \bibinfo{person}{Richard
  Souvenir}, {and} \bibinfo{person}{Robert Pless}.}
  \bibinfo{year}{2018}\natexlab{}.
\newblock \showarticletitle{Deep Randomized Ensembles for Metric Learning}. In
  \bibinfo{booktitle}{\emph{{ECCV} {(16)}}} \emph{(\bibinfo{series}{Lecture
  Notes in Computer Science})}, Vol.~\bibinfo{volume}{11220}.
  \bibinfo{publisher}{Springer}, \bibinfo{pages}{751--762}.
\newblock


\bibitem[\protect\citeauthoryear{Yu, Liu, Gong, Ding, and Tao}{Yu
  et~al\mbox{.}}{2018}]%
        {DBLP:conf/eccv/YuLGDT18}
\bibfield{author}{\bibinfo{person}{Baosheng Yu}, \bibinfo{person}{Tongliang
  Liu}, \bibinfo{person}{Mingming Gong}, \bibinfo{person}{Changxing Ding},
  {and} \bibinfo{person}{Dacheng Tao}.} \bibinfo{year}{2018}\natexlab{}.
\newblock \showarticletitle{Correcting the Triplet Selection Bias for Triplet
  Loss}. In \bibinfo{booktitle}{\emph{{ECCV} {(6)}}}
  \emph{(\bibinfo{series}{Lecture Notes in Computer Science})},
  Vol.~\bibinfo{volume}{11210}. \bibinfo{publisher}{Springer},
  \bibinfo{pages}{71--86}.
\newblock


\bibitem[\protect\citeauthoryear{Yuan, Yang, and Zhang}{Yuan
  et~al\mbox{.}}{2017}]%
        {DBLP:conf/iccv/YuanYZ17}
\bibfield{author}{\bibinfo{person}{Yuhui Yuan}, \bibinfo{person}{Kuiyuan Yang},
  {and} \bibinfo{person}{Chao Zhang}.} \bibinfo{year}{2017}\natexlab{}.
\newblock \showarticletitle{Hard-Aware Deeply Cascaded Embedding}. In
  \bibinfo{booktitle}{\emph{{ICCV}}}. \bibinfo{publisher}{{IEEE} Computer
  Society}, \bibinfo{pages}{814--823}.
\newblock


\bibitem[\protect\citeauthoryear{Zakizadeh, Sasdelli, Qian, and
  Vazquez}{Zakizadeh et~al\mbox{.}}{2018}]%
        {DBLP:journals/corr/abs-1807-11674}
\bibfield{author}{\bibinfo{person}{Roshanak Zakizadeh},
  \bibinfo{person}{Michele Sasdelli}, \bibinfo{person}{Yu Qian}, {and}
  \bibinfo{person}{Eduard Vazquez}.} \bibinfo{year}{2018}\natexlab{}.
\newblock \showarticletitle{Improving the Annotation of DeepFashion Images for
  Fine-grained Attribute Recognition}.
\newblock \bibinfo{journal}{\emph{CoRR}}  \bibinfo{volume}{abs/1807.11674}
  (\bibinfo{year}{2018}).
\newblock


\bibitem[\protect\citeauthoryear{Zhan, Shi, and Kot}{Zhan
  et~al\mbox{.}}{2017a}]%
        {DBLP:conf/bmvc/ZhanSK17}
\bibfield{author}{\bibinfo{person}{Huijing Zhan}, \bibinfo{person}{Boxin Shi},
  {and} \bibinfo{person}{Alex~C. Kot}.} \bibinfo{year}{2017}\natexlab{a}.
\newblock \showarticletitle{DeepShoe: {A} Multi-Task View-Invariant {CNN} for
  Street-to-Shop Shoe Retrieval}. In \bibinfo{booktitle}{\emph{{BMVC}}}.
  \bibinfo{publisher}{{BMVA} Press}.
\newblock


\bibitem[\protect\citeauthoryear{Zhan, Shi, and Kot}{Zhan
  et~al\mbox{.}}{2017b}]%
        {DBLP:conf/icip/ZhanSK17}
\bibfield{author}{\bibinfo{person}{Huijing Zhan}, \bibinfo{person}{Boxin Shi},
  {and} \bibinfo{person}{Alex~C. Kot}.} \bibinfo{year}{2017}\natexlab{b}.
\newblock \showarticletitle{Street-to-shop shoe retrieval with multi-scale
  viewpoint invariant triplet network}. In \bibinfo{booktitle}{\emph{{ICIP}}}.
  \bibinfo{publisher}{{IEEE}}, \bibinfo{pages}{1102--1106}.
\newblock


\end{thebibliography}

%
\appendix

\end{document}